\documentclass[letterpaper, 10 pt, conference]{ieeeconf}  

\IEEEoverridecommandlockouts                              

\usepackage{graphicx} 
\usepackage{pgfplots}
\usepackage{tikz}
\usetikzlibrary{pgfplots.units, shapes, arrows, positioning, calc}
\pgfplotsset{compat=1.18}
\usepackage{amsmath} 
\usepackage{amssymb}  
\usepackage{comment}
\usepackage{float}
\usepackage{booktabs}
\usepackage[hyphens]{url}

\title{\LARGE \bf Multi-agent Reinforcement Learning for Robotized Coral Reef Sample Collection}

\author{Daniel Correa$^{1}$, Tero Kaarlela$^{1}$, Jose Fuentes$^{1}$, \\ Paulo Padrao$^{1}$,  Alain Duran$^{2}$, and Leonardo Bobadilla$^{1}$
\thanks{$^{1}$Knight Foundation School of Computing and Information Sciences, Florida International University, Miami, Florida, USA.
        {\tt\small dcorr039@fiu.edu, tkaarlel@fiu.edu, plope113@fiu.edu, jfuen099@fiu.edu, bobadilla@cs.fiu.edu}}%
\thanks{$^{2}$ College of Arts, Sciences \& Education, Biological Sciences, Florida International University, Miami, Florida, USA. {\tt\small alduran@fiu.edu}}
}

\begin{document}

\maketitle
\thispagestyle{empty}
\pagestyle{empty}

\begin{abstract}
This paper presents a reinforcement learning (RL) environment for developing an autonomous underwater robotic coral sampling agent, a crucial coral reef conservation and research task. Using software-in-the-loop (SIL) and hardware-in-the-loop (HIL), an RL-trained artificial intelligence (AI) controller is developed using a digital twin (DT) in simulation and subsequently verified in physical experiments. An underwater motion capture (MOCAP) system provides real-time 3D position and orientation feedback during verification testing for precise synchronization between the digital and physical domains. A key novelty of this approach is the combined use of a general-purpose game engine for simulation, deep RL, and real-time underwater motion capture for an effective zero-shot sim-to-real strategy.







\end{abstract}

\section{INTRODUCTION}
\label{sec:introduction}

Coral reefs experience multiple stressors, including but not limited to bleaching events, diseases, pollution, and over-harvesting~\cite{Coral2003}. These stressors' simple or synergistic effects have driven coral abundance to levels where their future is uncertain without human interventions such as restoration programs. Preserving the coral reefs requires vital actions such as planting new corals, removing harmful debris, collecting coral samples for research, and monitoring underwater conditions~\cite{Yang2024,wally2024}. 

Scuba diving is thus far the primary field method of coral reef research and conservation in shallow water. The activity is limited by several factors, including intense training and certification process, logistical challenge and, most importantly, short bottom times dictated by decompression illness risks that increase with depth. As a result, researchers face limitations while collecting data and are restricted to shallow waters. These limitations explain the lack of information on mesophotic reefs, which have been considered sheltered ecosystems from climatic events \cite{slattery2024}.

Alternatively, remotely operated underwater vehicles (ROUVs) offer several benefits over human divers, such as extended bottom time, minimized human risk, integration with advanced sensor suites, and logistical ease of maintenance and deployment. 

However, ROUVs does have certain challenges for human teleoperators since they require constant human input, typically from a trained operator, and communication latency can make precise maneuvering difficult. From this arises a dilemma in which a ROUV operator can be subjected to stress, fatigue, and potentially functional impairment, thus diminishing the potential benefits. High-level autonomous control in ROUVs can ideally operate with reduced reliance on human operators and help to enhance efficiency for complex ROUV tasks~\cite{scharff2021}. 

Reinforcement learning can be used to train AI control models to control agents with near-human performance. One challenge is creating a realistic training environment and agent, a problem known as the sim-to-real gap. Another challenge is designing an RL strategy that aligns with the human-operator goals. Neither can be ignored when the goal is to transfer the learned policy to the real world. 

\begin{figure}[t]
      \centering
      \includegraphics[width=3in]{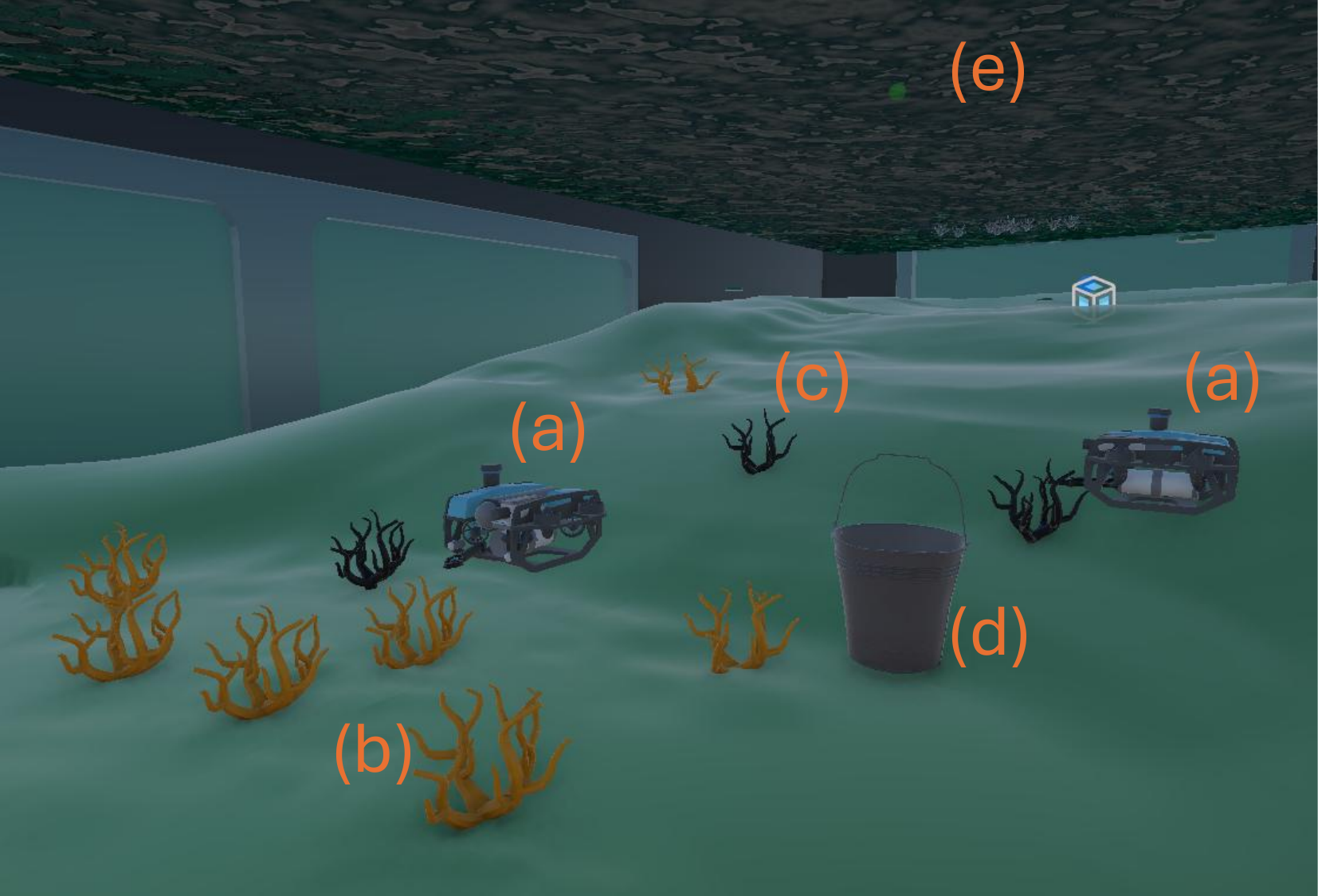}
      \caption{Illustration of the coral sample collection task, (a) Sample collecting vehicle, (b) healthy coral, (c) unhealthy coral, (d) sample collection container, (e) water surface.}
      \label{fig:introduction}
   \end{figure}

This paper introduces a RL environment for developing an AI controller for ROUVs. Its capability is demonstrated by transforming a low-cost ROUV into an autonomous underwater vehicle (AUV), which performs underwater searching and navigation tasks with reduced reliance on human operators, optimal efficiency, and the potential to scale up with multiple autonomous agents. The agent is trained to collect samples of healthy corals and to bring samples to a collection container, all while avoiding interaction with unhealthy corals. The environment is implemented using the Unity game engine with the dynamic water physics 2 package as well as the ML agents toolkit~\cite{juliani2020, DWS2}. The presented RL training environment is a digital replica of the underwater and surface vehicle testbed at Florida International University's Robotics and Autonomous Systems Laboratory for Coastal Conservation and Restoration (RASCAR). Figure~\ref{fig:introduction} illustrates the coral sample collection task.


The training algorithm incorporates a SIL/HIL methodology using the AI controller to command a digital twin of a ROUV in simulated training. Validation testing is done using the physical hardware. During validation testing, an underwater motion capture system (MOCAP) provides real-time position and orientation feedback. A key novelty of this approach is the combined use of a general-purpose game engine for simulation, deep RL, and real-time underwater motion capture for an effective zero-shot sim-to-real strategy.
The contributions of this paper are:

\begin{enumerate}
    \item Present an RL training environment to train an AI controller for an AUV search and navigation task. 
    \item Study the feasibility of the trained model in a zero-shot transfer case scenario.
\end{enumerate}

The rest of the paper is organized as follows: the next section introduces the previous research on autonomous control of underwater vehicles, Section~\ref{sec:materials} describes the materials used, Section~\ref{sec:methods} describes the methods of this study, Section~\ref{sec:results} details the results and Section~\ref{sec:discussion} provides the discussion, concluding the paper.


\section{RELATED WORK}
\label{sec:background}
The ability to monitor and manage complex marine environments is an ongoing challenge, and unmanned autonomous vehicles are more often being applied to help tackle this challenge~\cite{9389173}. Developing a reliable autonomous control method for AUVs is essential and is challenging due to the uncertainties present in unknown underwater environments~\cite{li2024}. Traditional control methods, such as Proportional-integral-derivative~\cite{jalving1994}, Sliding mode~\cite{qiao2020}, and Model predictive control~\cite{shen2018} are known to struggle with complex and uncertain hydrodynamic disturbances, noisy sensor data, and the need for robust real-time adaptation. Also, such general-purpose controllers are application agnostic and require significant effort to tailor to a specific application. 

To overcome these challenges, current research is focused on developing more advanced task-specific controllers for AUVs ~\cite{10610184} as well as AI-based control methods with fuzzy logic~\cite{Lakhekar2019}, Deep Neural networks (DNNs)~\cite{he2017}, and RL~\cite{huang2022, cai2024}. Such methods have been shown to provide effective autonomous control in multi-constraint underwater environments using simulators such as Gazebo~\cite{koenig2004}, MuJoCo~\cite{todorov2012}, and project DAVE~\cite{zhang2022dave}. However, a control model trained in a simulated training environment is typically not applicable in the real world, a problem known as the sim-to-real gap~\cite{TRENTSIOS2022287}. 
\subsection{Research Gap}
\label{subsec:background}

This work expands on previous work by leveraging the advanced capabilities of the Unity game engine as well as real-time underwater 3D motion capture to develop and verify an RL-based controller designed for a complex task ~\cite{10610184, cai2024}. We show the potential of this method as an effective zero-shot sim-to-real strategy for AUVs.

\section{MATERIALS}
\label{sec:materials}

\begin{figure}[!htbp]
      \centering
      \includegraphics[width=3in]{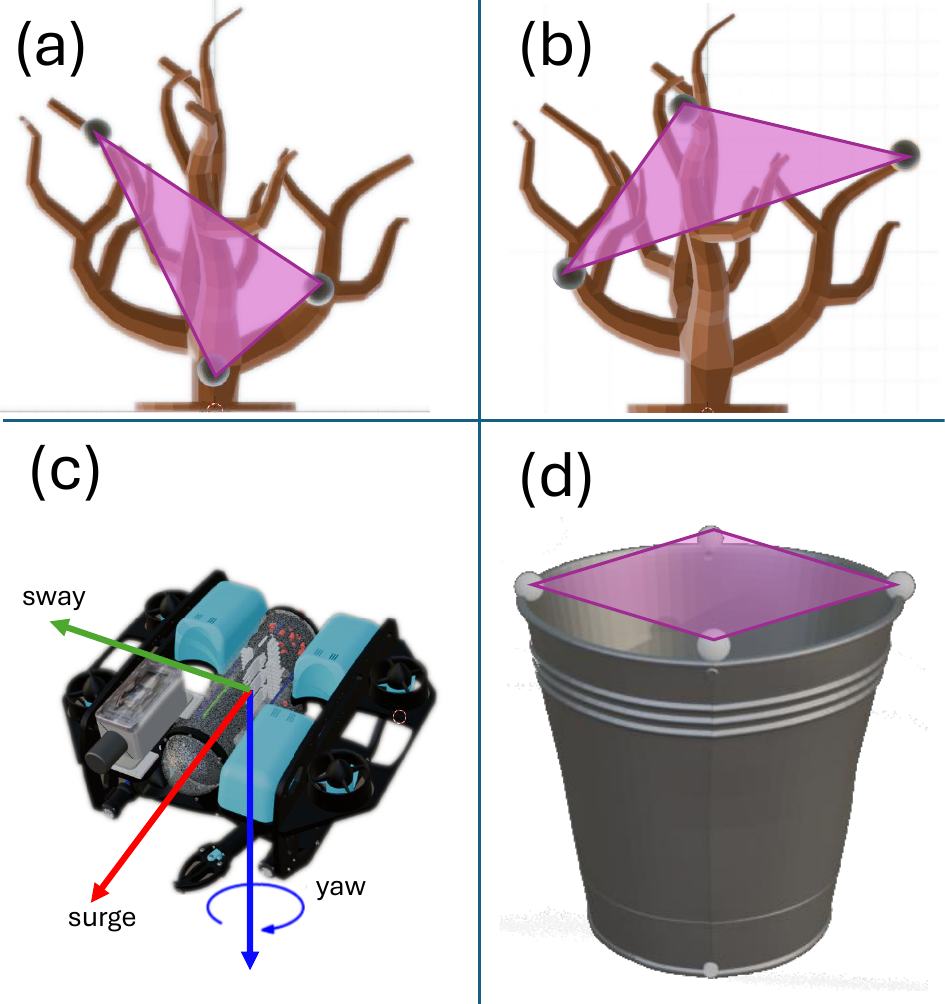}
      \caption{Materials utilized in this work (a) healthy coral mock-up, (b) unhealthy coral mock-up, (c) ROUV, (d) sample collection bucket.}
      \label{fig:materials}
\end{figure}

This section describes the materials used in this study. CAD models were used in simulations, while their physical counterparts were used during testing. Figures~\ref{fig:materials} and \ref{fig:architecture} illustrate the materials used in this study.\\

The RASCAR testbed includes a Qualisys underwater real-time motion capture system installed inside a 47-m$^3$ water tank. The motion capture system comprises nine cameras capable of tracking reflective markers; data is processed using Qualisys Track Manager (QTM) software. The calibrated capturing volume for underwater vehicles is approximately 6 × 3 × 1.5 m, with accuracy ranging from 1.5 to 3.0 mm. For this study, the motion capture frequency is set at 10 Hz. The corals used in this study are 3D-printed CAD models of the staghorn coral (\textit{Acropora cervicornis}). Healthy and unhealthy corals are distinguished by orange and white colors, respectively. The sample collection bucket is a standard 5-gallon bucket. The corals and bucket can be fitted with reflective markers to locate them in motion capture software.

The ROUV used in this study is a BlueROV2~\cite{bluerov2}. The BlueROV2 is a vectored 6-DoF vehicle designed for heavy payloads with eight thrusters and intrinsic orientation estimation and PID control capability. The ROUV has several sensory inputs, including an inertial navigation system (INS), depth sensor, forward-facing camera with a tilting mechanism, SONAR altimeter, and an acoustic communication modem. The ROUV is also equipped with an electrically actuated two-finger gripper. The INS provides attitude estimates as well as linear/angular velocity estimates relative to the body frame. The pressure sensor measures absolute pressure corresponding to the depth below the water's surface. The camera provides RGB images of the scene relative to the ROUV's pose. The SONAR altimeter measures vertical altitude relative to the surface below the ROUV. The acoustic modem functions as an omnidirectional ranging device between the ROUV and a collection bucket with the assumption that the collection bucket is also fitted with a complementary device. The range is simply the magnitude of distance. ROUV is controlled by open-source firmware and operating systems. A custom Python program based on the Python MAVLink (PyMAVlink) package was developed to interface the RL application with the ROUV control. Four reflective markers are attached in an asymmetrical pattern on the corners of the ROUV to capture its motion.

We use the Unity game engine to develop our virtual environment. The environment is a digital replica of Florida International University's underwater robotics testbed. 3D CAD models of the ROUV, 3D printed corals, and the collection bucket were used to populate the virtual environment. In Unity, we apply physical parameters (mass, center of mass, linear and angular damping, etc.) to each virtual object to simulate the rigid body mechanics of each given object accurately. Parameters for the virtual ROUV were determined based on physical measurements as well as values obtained from technical datasheets and published literature~\cite{Chu-JouWu-2018, cai2024, ROVdatasheet}. The \textit{Dynamic Water Physics 2} asset in Unity was used for hydrodynamics simulation~\cite{DWS2}. This package provides a platform for interaction between game objects and a simulated dynamic fluid field with customizable hydrodynamics parameters for different objects. The parameters for the virtual ROUV, such as buoyancy force, fluid density, skin drag coefficient, etc., were determined by referencing information from technical datasheets and published literature~\cite{Chu-JouWu-2018, cai2024, ROVdatasheet}.

\section{METHODS}
\label{sec:methods}

 In the following section, we describe the formulation of the coral collection problem in the context of RL, including the RL framework and algorithms used, the environment's features, and the agent's capabilities. We describe details about the network architecture, inputs, and outputs. Lastly, we describe the experimental design and the deliverables from each stage of the experiment.

\subsection{Problem Formulation and RL Algorithms}
\label{subsec:problemformulation}
The coral collection problem involves one or more ROUV agents that must explore an underwater environment to obtain healthy pieces of coral and deposit them in a collection bucket while avoiding unhealthy coral. We model the coral collection problem as a Markov Decision Process (MDP) with a standard formulation: $MDP = (S,A,P,R)$. $S$ indicates the state space, $A$ is the action space, $P$ indicates the transition probability function, $R$ represents the reward function. MDPs are iterative processes such that at every discrete time step $t$, the agent is in a current state $s_{t} \in S$ and receives a partial observation $o$ of its current state. Given $o$, the agent responds with an action based on its current policy $a_{t} =\pi_{\theta}(o)$. The agent subsequently obtains a reward $r_{t} = R(s_{t}, a_{t})$ and a new state $s_{t + 1} = P(s_{t}, a_{t})$. The training objective is to learn an optimal policy, $\pi_{\theta}$, which maximizes the expected cumulative long-term rewards. 

In this study, we apply methods from the Actor-Critic, a family of RL algorithms, to solve our problem. Namely, we apply the Proximal Policy Optimization (PPO) method~\cite{ppoalgo} and the Soft-Actor-Critic (SAC) method~\cite{sacalgo} in single-agent trials. The Independent PPO (IPPO) method~\cite{IPPO, yu2022} is also applied in a multi-agent trial. These methods are currently considered to be state-of-the-art algorithms and are well described in the RL corpus~\cite{Dutta_et_al.}. Also, the Unity ML-agents toolkit provides built-in support for these algorithms, allowing seamless integration into our application~\cite{juliani2020}.    

\subsection{Environment}
\label{subsec:environment}
Our environment is a compact 3D domain $\mathcal{W} \subseteq \mathbb{R}^{3}$; its deepest layer is a bathymetric surface $\mathcal{B} = \{(x,y,z) \in \mathcal{W}: z =\inf_{z'}\{z':(x,y,z') \in \mathcal{W}\}$ randomly populated with a set of coral objects $C$, which consists of two distinct disjoint subsets: healthy corals and diseased corals. The environment also contains a set of buckets $B$ randomly located within $\mathcal{B}$. The environment is assumed to have negligible hydrodynamic disturbances due to surface waves, while disturbances due to underwater currents are assumed to follow a linear pattern along a specific direction. While this general formulation applies to a wide variety of underwater landscapes in this study, we operate within the confines of the RASCAR testbed.\\

\subsection{ROUV Agent}
\label{subsec:rouv}
The agent is an ROUV model of a BlueROV2, incorporating all of its functionality. The agent's observation $o$ is a tuple with an RGB image from the forward-facing camera and a vector of scalar values from additional sensors. The observation vector can be described by $[\hat{u}, \hat{v}, \hat{r}, \hat{d}]^{T}$. $\hat{u}$, $\hat{v}$, and $\hat{r}$ correspond to estimated surge, sway, and yaw velocities from the onboard INS. $\hat{d}$ corresponds to estimated distances from the acoustic ranging device. Estimated values are obtained by adding random sensor noise to a modeled sensor output based on hardware specifications. The uncertainty of a given sensor can also be determined empirically for a given ROUV. 

The ROUV maneuvers according to the kinematic and kinetic models established in ~\cite{7393951}: 
\begin{equation}
    \begin{aligned}
        \dot{\eta} &=J(\eta)\nu \\
        \tau + w &= M\dot{\nu} + C(\nu)\nu + D(\nu)\nu + g(\eta)
        \\
    \end{aligned}
\end{equation}

where $\eta$ is the vehicle's pose in the world frame, $J(\eta)$ is the transformation matrix from the body frame into the world frame, $\nu$ is the vehicle's velocities relative to the body frame, $\tau$ is propulsion forces and moments acting on the vehicle, $w$ is the vector of unmodeled disturbances, $M$ is the inertia matrix, $C(\nu)$ is the Coriolis and centripetal forces matrix, $D(\nu)$ is the hydrodynamic damping matrix, and $g$ is the gravitational and buoyancy forces.

The agent is designed to take advantage of the lower-level PID controller by deferring control of some DOF of motion to the PID. Specifically, the ROUV is assumed to utilize PID control for stabilizing roll and pitch orientation angles relative to normal gravity. As well, the ROUV utilizes terrain following PID control to maintain a safe altitude above $\mathcal{B}$. With these considerations, the agent's action space $A=[-1,1]^3$ contains the motion controls $(u, v, r)\in A$, where $u$, $v$, and $r$ are analog control inputs setting the vehicle's surge, sway, and yaw velocities, respectively.

\subsection{Reward Shaping}
\label{subsec:reward}
\begin{figure}[!htbp]
      \centering
      \begin{tikzpicture}[node distance=1cm, every node/.style={align=center}]

        \tikzstyle{state} = [rectangle, draw=black, thick, minimum width=2.0cm, minimum height=1.5cm]
        \tikzstyle{good} = [state, fill=green!80, draw=green!80, text=white] 
        \tikzstyle{bad} = [state, fill=red!80, draw=red!80, text=white] 
        \tikzstyle{bucket} = [state, fill=blue!80, draw=blue!80, text=white] 
        \tikzstyle{searching} = [state, fill=white]
        
        \node[bucket] (bucket) {Finds \\ Bucket};
        \node[searching, above=of bucket] (coral) {Coral \\ Searching};
        \node[good, left=of bucket] (good) {Finds \\ Good Coral};
        \node[bad, right=of bucket] (bad) {Finds \\ Bad Coral};
        \node[searching, below=of bucket] (bucket_search) {Bucket \\ Searching};
        
        \draw[thick, ->] (coral) edge[loop above] (coral);
        \draw[thick, ->] (coral.west) to[bend right=10] node[above left] {$R = 1$ \\ $G = True$} (good.north);
        \draw[thick, ->] ($(coral.east)!0.5!(coral.north east)$) to[bend left=10] node[above right] {$R = -1$} ($(bad.north)!0.5!(bad.north east)$) ;
        \draw[thick, ->] ($(bad.north)!0.5!(bad.north west)$) to[bend left=10] node[below] {$G is false$} ($(coral.east)!0.5!(coral.south east)$);
        \draw[thick, ->] ($(coral.south)!0.5!(coral.south west)$) to[bend right=10] node[left] {R = -0.1} ($(bucket.north)!0.5!(bucket.north west)$);
        \draw[thick, ->] ($(bucket.north)!0.5!(bucket.north east)$) to[bend right=10] node[above right] {} ($(coral.south)!0.5!(coral.south east)$);
        \draw[thick, ->] ($(bad.south)!0.5!(bad.south west)$) to[bend right=10] node[below right] {G is true} ($(bucket_search.north east)!0.5!(bucket_search.east)$);
        \draw[thick, ->] ($(bucket_search.south east)!0.5!(bucket_search.east)$) to[bend right=10] node[below right] {R = -1}   ($(bad.south east)!0.5!(bad.south)$);
        \draw[thick, ->] ($(good.south)!0.5!(good.south west)$) to[bend right=10] node[left] {$R = -0.1$} ($(bucket_search.west)!0.5!(bucket_search.south west)$);
        \draw[thick, ->] ($(bucket_search.west)!0.5!(bucket_search.north west)$) to[bend right=10] node[left] {} ($(good.south)!0.5!(good.south east)$);
        \draw[thick, ->] (bucket_search) -- node[left] {$R = 1$ \\ $G = False$} (bucket);
        \draw[thick, ->] (bucket_search) edge[loop below] node[below] {$R =  \ln(1 + \text{bucket distance})$} (bucket_search);
        
    \end{tikzpicture}
    \caption{Reward model for the coral collection agent}
      \label{fig:rewardmodel}
\end{figure}
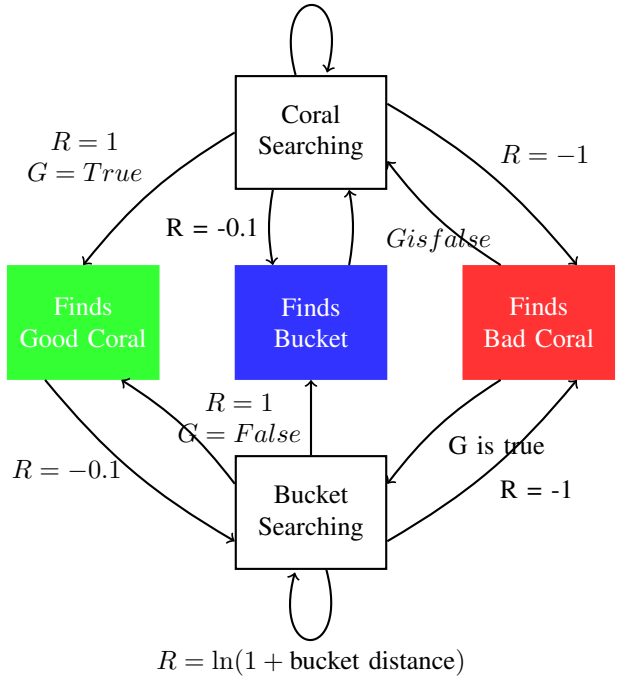

The reward model of the coral collection agent is simply a Deterministic Finite State Machine (DFSM) resembling a policy for collecting coral. Figure~\ref{fig:rewardmodel} illustrates the reward model used.

The reward structure is conducive to learning by using elements of the agent's perceptible environment to promote the desired actions. An optimal policy maximizes the cumulative reward to the fullest extent. In Figure~\ref{fig:rewardmodel}, $R$ represents a single reward received from a given state transition, and $G$ is a single memory element that keeps track of whether or not the agent has obtained a coral sample. 

\subsection{Network Architecture}
\label{subsec:architecture}

\begin{figure}[!htbp]
      \centering
      \includegraphics[width=3in]{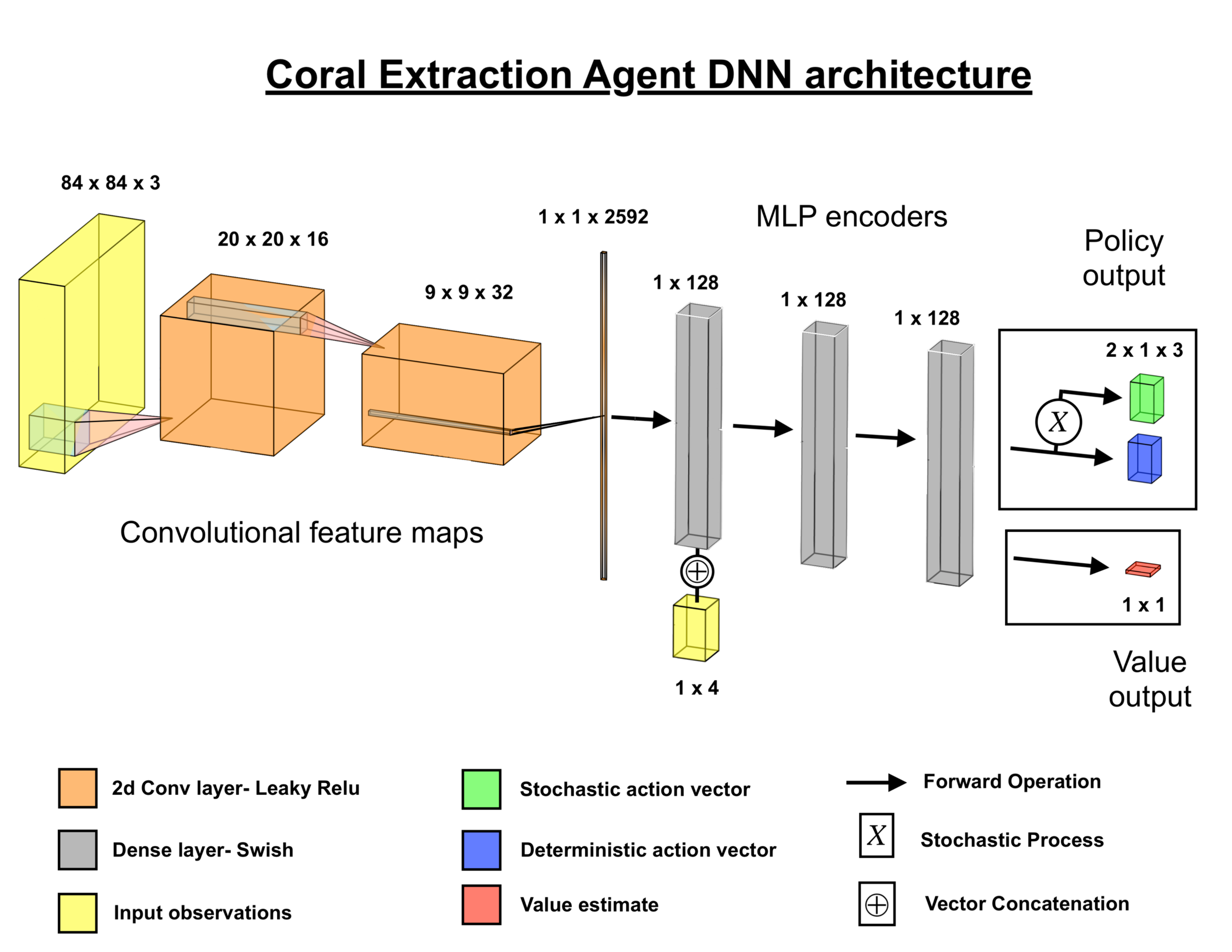}
      \caption{Network architecture used in the PPO experiment. A Critic network outputs a value estimate while an Actor network outputs action vectors.}
      \label{fig:nnarchitecture}
\end{figure}

\begin{figure*}[!htb]
  \centering
  \includegraphics[width=\textwidth]{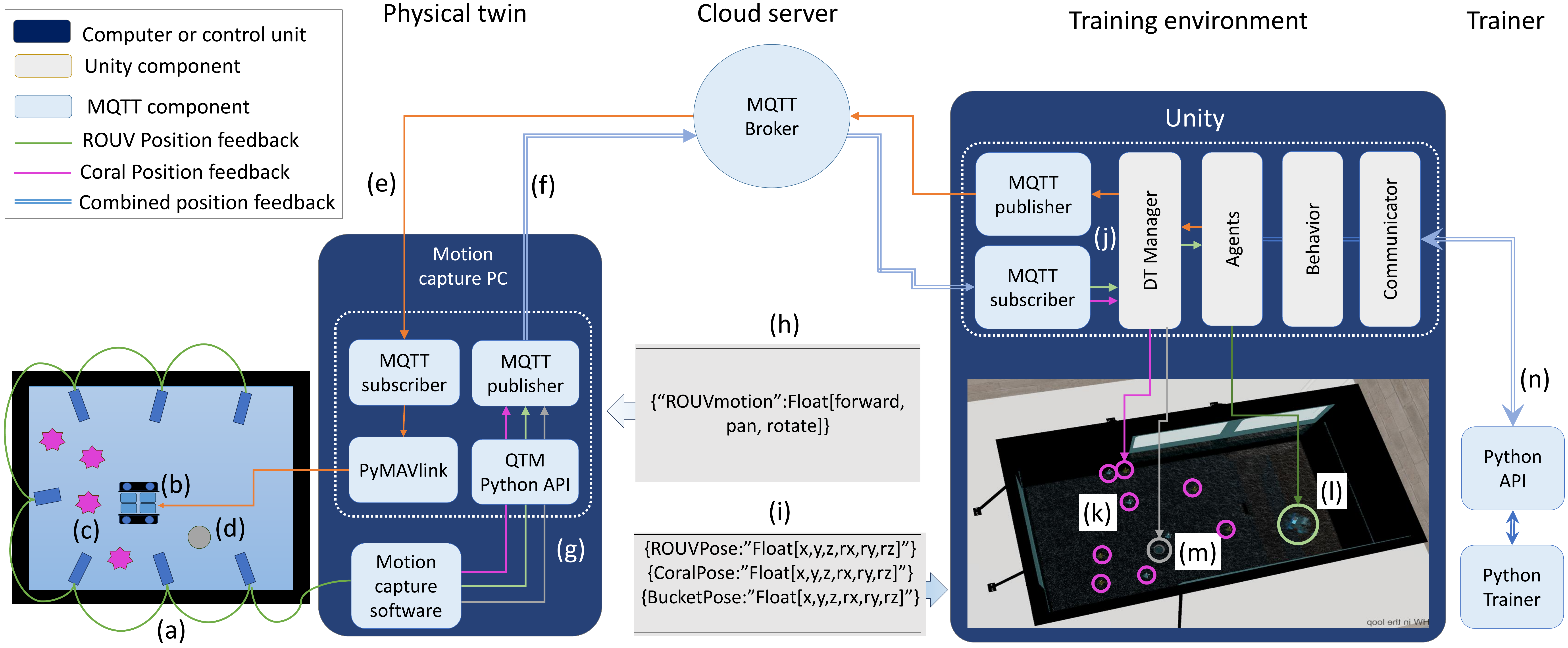}
  \caption{The testing environment HIL architecture. (a) The underwater motion capture cameras, (b) ROUV, (c) Coral mockups, (d) Sample collection bucket, (e) ROUV motion commands, (f) Combined ROUV, coral, and sample bucket position feedback, (g) custom Python application to interface with the physical environment, (h) message structure to command the ROUV motion, (i) message structure for combined position feedback, (j) Unity software modules to interact with the physical and digital environments, (k) positions of the detected corals, (l) position of the agent, (m) position of the sample bucket, (n) ML-Agents Python trainer.}
  \label{fig:architecture}
\end{figure*}

The agent's critic and actor functions are DNN architectures designed to estimate either values or actions using the multi-modal inputs from $o$. A vectorized deep convolutional neural network (CNN) is fitted with a multilayer perceptron (MLP) head to predict a given output. The vector observation of $o$ is concatenated at the base of the MLP to provide this observation weight. This method allows image observations to be independently weighed against vector observations. We used a consistent DNN architecture in all experimental trials. Figure~\ref{fig:nnarchitecture} graphically depicts the network architecture used in the PPO experiment.

\subsection{Experimental Design}
\label{subsec:experiment}

We performed a set of experiments in which each RL algorithm is applied with a randomly initialized agent and environment. Training is entirely simulated and the AI controller is trained in a SIL approach. Each agent is trained for up to $8\times{10^{6}}$ steps using default hyper parameters for each algorithm. Table 1 details several variables controlled in each trail of the experiment:  

\begin{table}[!htb]
    \centering
    \caption{Table of independent variables for each experimental trail}
    \label{tab:motionvalues}
    \begin{tabular}{|c||c||c||c|} \hline 
         Trial / Variables&  \# of agents& \# of good/bad corals& \# of buckets\\ \hline 
         PPO&  1&  5& 1\\ \hline 
         SAC&  1&  5& 1\\ \hline
         IPPO& 3& 15& 3\\ \hline
    \end{tabular}    
\end{table}

The result of each experiment at this stage is a trained AI controller model. Based on the training results, we then choose the best performing model to subsequently be used for validation testing.

 The physical environment consists of one ROUV agent, one healthy coral, and one bucket. In this environment, the physical counterpart of each object is present, and the 3D pose of each object is measured using real-time underwater MOCAP data. This real-time data is applied within Unity to affect the 3D pose of the digital twins. Thus, the agent's 3D pose in the testing environment is not estimated using simulation as in training but is instead obtained using ground truth data. Using only inference, the AI controller signals are directed to the physical ROUV agent rather than the digital twin. Thus, we effectively emulate the ROUV agent using HIL. Figure 4 illustrates the architecture of the cyber-physical system used to enable HIL testing.


\section{RESULTS}
\label{sec:results}

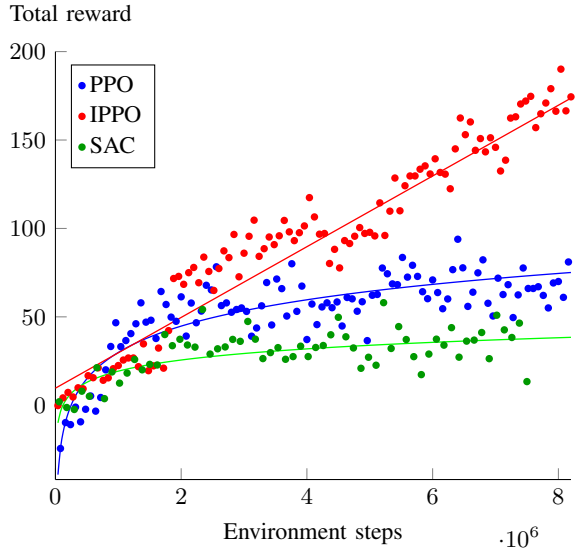
\begin{figure}[!htb]
    \centering
        \centering
        \begin{tikzpicture}
        \pgfplotsset{every tick label/.append style={font=\small}}
        \pgfplotsset{xlabel/.append style={font=\small}}
        \pgfplotsset{every non boxed y axis/.append style={y axis line style=-}}
        \pgfplotsset{every axis plot/.append style={thick}}
            \begin{axis}[
                title={},
                mark repeat={2},
                legend cell align={left},
                xmin=-1, xmax=8200000,
                ymin=-42, ymax=200,
                ycomb,             
                xlabel={Environment steps},
                ylabel={Total reward},  
                ylabel style={at={(axis description cs:0.03,1.05)}, anchor=south,rotate=-90},  
                ylabel shift={10pt},  
                x tick label style={/pgf/number format/1000 sep=},
                scaled y ticks = false,
                scaled x ticks = true,
                y tick label style={/pgf/number format/fixed},
                axis x line*=bottom,
                y axis line style={draw opacity=30},
                axis y line*=left,
                x axis line style={draw opacity=30},
                legend pos=north west,
            ]

            \addplot[color=blue,only marks,mark color=blue,fill=blue,mark size=1pt] table [col sep=comma]   {Datasets/reward.tex}; 
            \addplot[color=red,only marks,mark color=red,fill=red, mark size=1pt] table [col sep=comma]   {Datasets/large_scale_reward.tex};
            \addplot[color=green!60!black,only marks,mark color=green!60!black,fill=green!60!black, mark size=1pt] table [col sep=comma]   {Datasets/sac_reward.tex};
            \addlegendentry{PPO}; 
            \addlegendentry{IPPO};
            \addlegendentry{SAC};
            \addplot[smooth, black, line width=0.5pt,domain=-42:8200000, samples=205, color=blue]  {21.49*ln(x) - 267.03}; 
            \addplot[smooth, black, line width=0.5pt,domain=-42:8200000, samples=205, color=red, no marks] {2e-5*(x) + 9.67};
            \addplot[smooth, black, line width=0.5pt,domain=-42:8200000, samples=205, color=green]  {9.09*ln(x) - 106.29}; 

            
        \end{axis}
        \end{tikzpicture}
    \caption{Episode rewards over environment steps using PPO, SAC, and IPPO training algorithms.}
    \label{fig:reward}
\end{figure}

During the training phase, the accumulated rewards of each agent were recorded. Figure~\ref{fig:reward} shows the time series of accumulated rewards. 

It can be seen that by the end of the training regimen the IPPO experiment has accumulated the most rewards, while the PPO experiment has overtaken the SAC experiment in terms of accumulated rewards.  

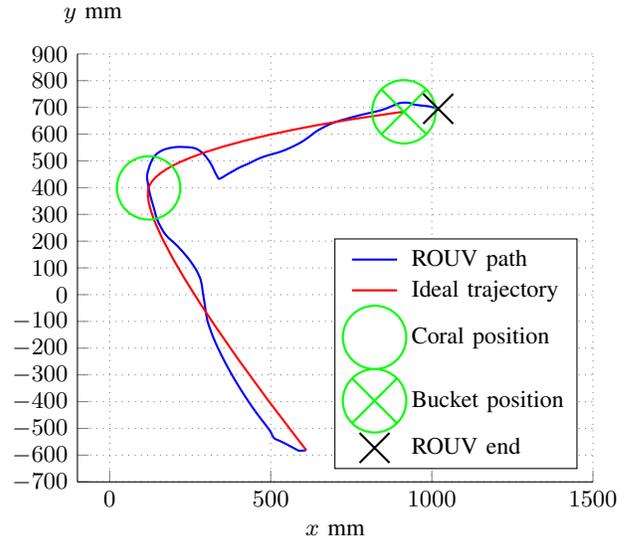
\begin{figure}[!htb]
    \centering
        \centering
        \begin{tikzpicture}
        \pgfplotsset{every tick label/.append style={font=\small}}
        \pgfplotsset{xlabel/.append style={font=\small}}
        \pgfplotsset{every non boxed y axis/.append style={y axis line style=-}}
        \pgfplotsset{every axis plot/.append style={thick}}
            \begin{axis}[
                title={},
                mark repeat={2},
                legend cell align={left},
                xmin=-100, xmax=1500,
                ymin=-700, ymax=900,
                ycomb,             
                xlabel={$x$ mm},
                ylabel={$y$ mm},  
                ylabel style={at={(axis description cs:0.03,1.05)}, anchor=south,rotate=-90},  
                ylabel shift={10pt},  
                x tick label style={/pgf/number format/1000 sep=},
                scaled y ticks = false,
                scaled x ticks = true,
                y tick label style={/pgf/number format/fixed},
                axis x line*=bottom,
                y axis line style={draw opacity=30},
                axis y line*=left,
                x axis line style={draw opacity=30},
                legend pos=south east,
                ytick distance=100,
                xtick distance=500,
                grid=both, 
                grid style={dotted, gray},
            ]

            \addplot[color=blue, smooth, thick] table [col sep=comma] {Datasets/rovpose_set4.tex};
           \addlegendentry{ROUV path}
            \addplot[color=red, smooth, thick] coordinates {(610.291589,-581.087432)(120.898126,399.113430)(913.111808,683.606202)};
            \addlegendentry{Ideal trajectory}
            \addplot[color=green,only marks,mark color=green,fill=green,mark size=16pt, mark=o, mark options={scale=0.75}] coordinates {(120.898126,399.113430)};
            \addlegendentry{Coral position}
            \addplot[color=green,only marks,mark color=green,fill=green,mark size=16pt, mark=otimes, mark options={scale=0.75}] coordinates {(913.111808,683.606202)};
            \addlegendentry{Bucket position}
            \addplot[color=black,only marks,mark color=black,fill=black,mark size=8pt, mark=x] coordinates {(1019.691799,694.8003143)};
            \addlegendentry{ROUV end}
 
        \end{axis}
        \end{tikzpicture}
    \caption{ROUV trajectory compared to an ideal trajectory.}
    \label{fig:feasibility}
\end{figure}

During the testing phase, the 3D pose of the agent was also recorded. Figure~\ref{fig:feasibility} depicts the pose data relative to the (x, y) plane of the testing environment. The plot shows the locations of the coral and bucket with an accuracy relative to $\pm$ 100mm.

\begin{figure}[!htbp]
      \centering
      \includegraphics[width=3in]{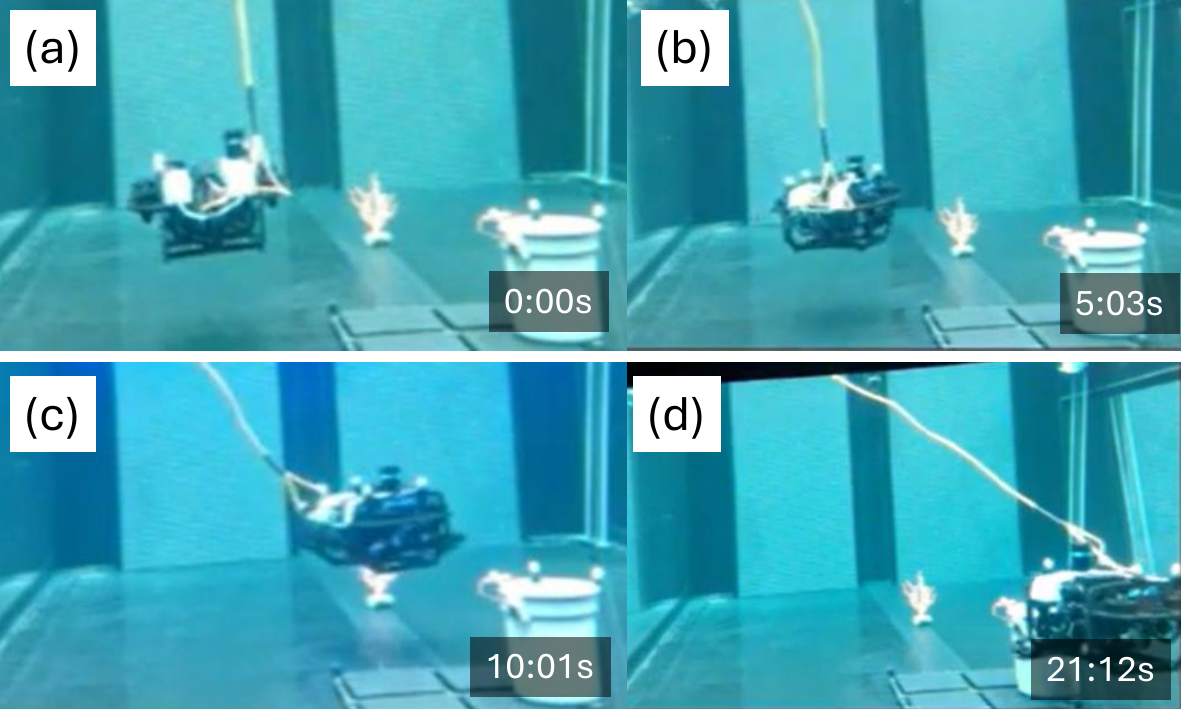}
      \caption{The coral sample collection task: (a) starting point, (b) detection of the coral, (c) coral collection, (d) sample drop.}
      \label{fig:timeseries}
\end{figure}

Figure~\ref{fig:timeseries} shows video stills from real-time footage of the HIL testing. Each image depicts key aspects of the agent's behavior and the time stamp of each given moment.

\section{DISCUSSION}
\label{sec:discussion}

The training results show us that each agent does learn an advantageous policy. This is evident by the characteristic logarithmic growth functions present in the training curves for both the PPO and SAC agents. Such a trend shows that both agents learned to maximize their utility in the training environment. The IPPO training result shows that multi-agent system also had positive learning outcomes. In fact, based on the seemingly linear training trend, it can be inferred that there was still more room for improvement. However, this could only be verified with longer training times. By the end of the prescribed training process, the IPPO system was accumulating about 3 times as many rewards as the single agent environment. This is consistent with the fact that the IPPO system was trained with a factor of 3 times as many environmental features and agents. These results indicate that the RL framework can be successfully scaled up with multiple agents offering a potentially linear increase in effectiveness. 

We hypothesize that simplifying the RL framework by reducing the agent's action space, as well as the potential complexity of the state space, benefits learning. For instance, under these assumptions, we considered the agent only needs to learn to manipulate 3-DOF of motion, specifically surge, sway, and yaw velocities, thus reducing the potential action space to half while still allowing complete navigational autonomy across $\mathcal{B}$. Furthermore, the terrain-following controller will restrict the robot's operation to $\mathcal{B}$, simplifying the potential state space by focusing the agent's effort on the most important component of the state space. We believe that such simplifications fully reconcile the navigation challenges of this problem with a synergy between the classical PID controller and the RL-based AI controller. This hypothesis could be tested by attempting to train the same agent with the full 6 DOF of motion, allowing the agent to explore shallower regions of the water column and/or potentially collide with the bathymetric surface.

Our analysis of the testing data is compared with an ideal trajectory. We believe the results demonstrate the feasibility of the trained model. We observe that the deviations from an ideal trajectory are usually less than 70 mm. Considering the relatively short training time, the results are favorable for a zero-shot transfer case scenario.

We believe the presented work is a step closer to having an ideal DT of the RASCAR lab and the ROUV. Optimizing the parameters of the DT will be a focus of future work. It is worth noting that the training environment used in this work represents optimal conditions that only require vehicle and water parameters, which have been determined in published research. Finding the optimal parameters to achieve the realistic behavior of objects in a simulated environment is challenging given the complexity of the physical mechanics involved and time-consuming if such parameters need to be determined empirically.

We plan to expand on this work directly by incorporating AI-image recognition and gripper functionality into the ROUV so that it can effectively emulate the coral collection and deposit tasks described in this study. In the future, an open sea training environment could be developed where the vehicle model is subjected to disturbances caused by sea currents, and the environment features many forms of marine life. For the open sea validation experiments, the feedback from the QTM motion capture system can be replaced by any system capable of providing $x$, $y$, and $z$ coordinate feedback, such as an underwater GPS system~\cite{bluerovgps}.

\section*{ACKNOWLEDGMENT}
This work is supported in part by NSF grants IIS-2024733 and IIS-2331908, the Office of Naval Research grant N00014-23-1-2789, the U.S. Department of Homeland Security grant 23STSLA00016-01-00, the U.S. Department of Defense grant 78170-RT-REP, and Florida Department of Environmental Protection grant INV31. 

\bibliographystyle{IEEEtran}
\bibliography{root}
\end{document}